# A Patient-Specific Digital Twin for Adaptive Radiotherapy of Non-Small Cell Lung Cancer

**Anvi Sud[1], Jialu Huang[2], Gregory R. Hart[3], Keshav Saxena[4], John Kim[5], Lauren Tressel[5], Jun Deng[5,6*]**

[1]Computational Biology and Biomedical Informatics Program, Yale University, New Haven, CT, USA

[2]Department of Statistics and Data Science, Yale University, New Haven, CT, USA

[3]Division of Natural Sciences, University of Guam, Mangilao, GU, USA

[4]Department of Biology and Computer Science, University of North Carolina at Chapel Hill, NC, USA

[5]Department of Therapeutic Radiology, Yale University, New Haven, CT, USA

[6]Department of Biomedical Informatics and Data Science, Yale University New Haven, CT, USA

\* Correspondence:
Jun Deng, PhD
jun.deng@yale.edu

**Abstract**

Radiotherapy continues to become more precise and data dense, with current treatment regimens generating high-frequency imaging and dosimetry streams ideally suited for AI-driven temporal modeling to characterize how normal tissues evolve with time. Each fraction in biologically guided radiotherapy (BGRT)-treated non-small cell lung cancer (NSCLC) patients records new metabolic, anatomical, and dose information. However, clinical decision-making is largely informed by static, population-based NTCP models which overlook the dynamic, unique biological trajectories encoded in sequential data. Even though normal-tissue response fluctuates mid-course through transient dose redistributions, metabolic shifts, or subtle radiomic changes that precede symptoms by days, most toxicity prediction models using machine learning yield a single endpoint estimate.

We developed COMPASS (COMprehensive Personalized ASsessment System) for safe radiotherapy functioning as a temporal digital-twin architecture utilizing per-fraction PET/CT, dosiomics, radiomics, and cumulative biologically equivalent dose (BED) kinetics to model normal-tissue biology as a dynamic time-series process. A GRU autoencoder was employed to learn organ-specific latent trajectories, which were classified via logistic regression to predict eventual CTCAE ≥1 toxicity.

Eight NSCLC patients undergoing BGRT contributed to the 99 organ-fraction observations that covered 24 organ trajectories (spinal cord, heart, and esophagus). Despite the small cohort, intensive temporal phenotyping allowed for comprehensive analysis of individual dose-response dynamics. In a leave-one-out evaluation, COMPASS had an AUC of 0.90, 80% sensitivity, and 78.6% specificity. Our findings revealed a viable AI-driven early-warning window, as increasing risk ratings occurred from several fractions before clinical toxicity. The dense BED-driven representation revealed biologically relevant spatial dose-texture characteristics that occur before toxicity and are averaged out with traditional volume-based dosimetry.

COMPASS establishes a proof of concept for AI-enabled adaptive radiotherapy, where treatment is guided by a continually updated digital twin that tracks each patient's evolving biological response.

## 1 Introduction

Locally advanced non-small cell lung cancer (NSCLC) radiation therapy represents a core therapeutic dilemma: escalating tumor dose improves local tumor control but increases the risk of severe normal tissue toxicity to adjacent critical structures such as the heart, lungs, esophagus and spinal cord. Current clinical practices rely on population-derived normal tissue complication probability (NTCP) models and dose–volume histogram (DVH) constraints to evaluate toxicity risk during treatment planning[1]. While these techniques have guided safe therapy for decades, they face critical limitations in the era of adaptive, biology-guided radiotherapy. Anatomical deformation, organ mobility, tumor regression and the evolving biological response that occurs during fractionated delivery cannot be accounted through static treatment evaluations based on planning CT [2]. Standard DVH metrics, while capturing volumetric dose distributions, aggregate dose across entire organs without preserving spatial heterogeneity or integrating voxel-wise biological markers that reflect early injury signals[3]. Above all, these methodologies do not provide a concrete mechanism for updating risk predictions in real time as new imaging becomes available at each treatment fraction thereby limiting the possibility of an adaptive intervention before toxicity becomes clinically apparent[4].

The procedure for radiotherapy inherently presents high frequency longitudinal data that remains largely underutilized with each fraction generating a new PET/CT scan. These streams record geometric changes through updated anatomical contours such as evolving tissue metabolism and delivered dose distribution reflecting plan-delivery fidelity. These temporal datasets encode patient-specific dose-response trajectories including cumulative dose kinetics, early metabolic abnormalities, evolving radiomic signatures, and dynamic organ sensitivity patterns. When modeled appropriately, such data offers the possibility to quantify not only organ-level toxicity risk but also per-voxel spatial susceptibility, enabling visualization of regions close to biological thresholds. By integrating voxel-wise biologically equivalent dose (BED) accumulation, organ-specific thresholds, and spatiotemporal dose heterogeneity, one can derive early spatial indicators of where injury may emerge during treatment to enhance conventional planning[5].

Recent advances in sequential deep learning, multimodal representation learning, and physics-informed modeling provide an opportunity to reconceptualize radiotherapy that treats treatment response as a dynamic biological process rather than a single pre-treatment snapshot. While large retrospective cohorts provide power for estimating population trends, they typically lack longitudinal imaging and dosimetry. On the other hand, per-fraction data collection in prospective radiotherapy settings yields detailed temporal information within each patient but necessarily limits cohort size [6]. We argue that this shift from broad sampling across many individuals to deep phenotyping within each patient is not a statistical compromise, but a design principle well aligned with biologically guided radiotherapy. Comprehensive longitudinal characterization enables models to track subtle biological deviations that would be averaged out in cross-sectional datasets, like how climate and weather systems rely on dense, continuous sensor data to track dynamic environmental processes.

To put this concept into practice, we developed COMPASS, a patient-specific temporal framework that integrates multimodal imaging features, organ-level dosimetric descriptors, and recurrent neural architectures to estimate the evolving normal tissue toxicity risk throughout fractionated radiotherapy[7]. The approach blends unsupervised representation learning through GRU(Gated Recurrent Network) autoencoding with supervised toxicity classification, enabling extraction of compact temporal embeddings that reflect each organ's biological trajectory[8,9]. By incorporating physics derived equations with cumulative dose distributions and organ thresholds, we were further able to per-voxel toxicity heatmaps allowing spatial localization of emerging hotspots of biological vulnerability. The framework addresses two core research questions: Do individual dose-response trajectories contain early biological signals preceding clinical toxicity manifestation? Can rich temporal phenotyping compensate for limited cohort size and allow patient-specific digital twins to generalize to individuals with atypical dosage patterns?

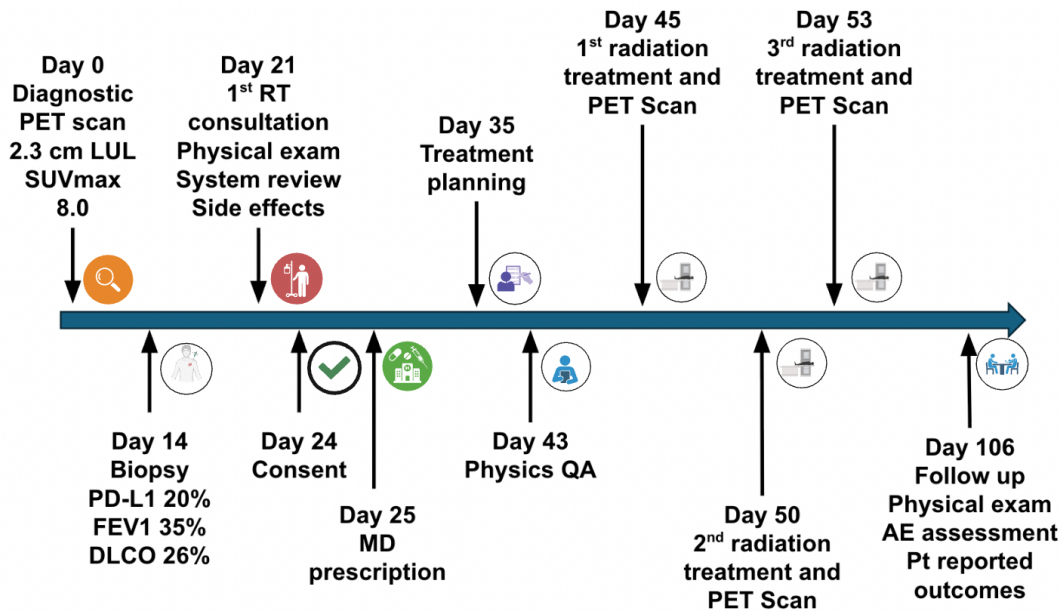

**Figure 1**. BGRT Workflow

## 2    Methods

### 2.1    Patient Cohort and Data Acquisition

Eight patients with locally advanced non-small cell lung cancer receiving definitive fractionated radiotherapy were included in this proof-of-concept study[10]. Treatment regimens ranged from 3 to 5 fractions per patient, generating 99 total organ-fraction observations across 24 individual organ trajectories (heart, esophagus, spinal cord). For each treatment fraction, longitudinal PET/CT imaging was acquired. Planned and delivered dose information were used to create per-fraction dose distributions to support fraction-specific analysis and cumulative tracking.

Toxicity outcomes were prospectively graded according to the Common Terminology Criteria for Adverse Events (CTCAE) version 5.0 within 6-11 follow up months of treatment completion[11]. With **10 out of 24 organ trajectories** exhibiting toxicity, binary classification categorized results as Grade 0 (no toxicity) against Grade ≥1 (any evident harm). This ratio reflected the high-risk nature of thoracic radiotherapy near dose-limiting organs while providing sufficient outcome diversity for supervised learning.

## 2.2 Data Preprocessing and Region of Interest (ROI) Delineation

To ensure consistency across the integration of multimodal features at the voxel level, we implemented a standardized preprocessing pipeline that co-registered longitudinal PET, kVCT, RTDose, and RTSTRUCT data into a single, patient-specific anatomical coordinate system[12]. This processing was performed utilizing core libraries including pydicom (version 2.4.4) for DICOM parsing, SimpleITK(2.5.0) for image manipulation and registration, and nibabel for NIfTI conversion, a combination of tools recognized for constructing robust medical image processing workflows [13,14,15].

For each patient, the time-averaged Average Intensity Projection (AIP) CT scan served as the spatial reference frame for all longitudinal data[16]. This AIP CT was reconstructed from the complete per-fraction CT series using `SimpleITK.ImageSeriesReader()` and exported as a NIfTI volume (typical size: 512 × 512 × 175 voxels). Regions of interest (ROIs) for the heart, esophagus, and spinal cord were generated by parsing the `ROIContourSequence` from RTSTRUCT files and converting the DICOM contours into 3D binary masks using rt-utils which were then saved as NifTi using SimpleITK[17]. Geometric transforms derived from DICOM metadata were applied to ensure exact co-registration of these ROI masks with the AIP reference frame. The final AIP-aligned dataset provided a unified voxel-wise framework, facilitating the extraction of synchronized radiomic, dosiomic, and temporal features from per-fraction PET/kVCT images, delivered dose distributions, and cumulative BED maps across all treatment fractions[18, 19].

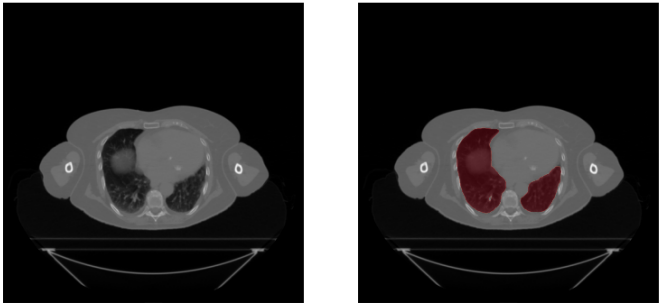

**Figure 2.** Visualization of the Average Intensity Projection (AIP) CT scan (slice 89) serving as the spatial reference frame, shown alongside the Lung segmentation which is displayed as a Region of Interest (ROI) overlay on the CT image.

### 2.3 Multimodal Feature Engineering

Using a voxel-wise biological dose computation pipeline intended to maintain spatial dose variability, we retrieved 73 characteristics from seven complementary modalities for every patient-organ-fraction observation shown in Table 1. We first computed per-voxel biological effective dose (BED) for each fraction using the linear-quadratic model BED = d × (1 + d/(α/β)), where d is the physical dose per voxel and α/β was organ-specific (3 Gy for heart and esophagus and 2 Gy for spinal cord). Voxel wise BED values were then summed across fractions to yield cumulative BED per voxel, accounting for the nonlinear dose-response where high-dose voxels contribute disproportionately to biological effect. This cumulative voxelwise BED distribution was converted to equivalent dose in 2-Gy fractions (EQD2) via EQD2 = BED / (1 + 2/(α/β)), preserving spatial heterogeneity before feature aggregation[5]. From these voxel wise EQD2 distributions, we extracted traditional dosimetric features comprising dose statistics (Dmax, Dmean, Dmin, D5, D50, D95), volume-threshold metrics (D2cc, V5, V10, V20), and plan quality indices (dose homogeneity, conformity, and gradient indices)[20]. Critically, dose distribution features performed spatial analysis of per-fraction physical dose distributions (prior to cumulative summation), extracting percentile doses (D2p through D98p quantifying dose to top x% of voxels), volume-percentile metrics (V10pMax through V90pMax indicating percentage receiving >x% of maximum dose), and distribution shape parameters (skewness, kurtosis)[21,22]. These per-fraction dosiomics capture intrafraction dose heterogeneity patterns, while cumulative EQD2 features quantify total biological burden; together they provide complementary temporal (per-fraction patterns) and cumulative (biological endpoint-relevant) dose characterization. The percentile-based dosiomics demonstrated superior sensitivity to focal hotspots compared to traditional volume thresholds in the results; for instance, D2p detects 10 Gy concentrated in 2% of the spinal cord (potentially myelopathic) despite V10=0%, a critical advantage for serial organ toxicity prediction.

Temporal kinetics features tracked dose accumulation dynamics via change in cumulative mean EQD2 per fraction (ΔDmean_EQD2) and inter fraction time intervals, capturing recovery between treatments. Geometric features quantified organ volume and fractional hotspot volume (percentage receiving >80% of Dmax). CT and PET intensity radiomics extracted first-order statistical descriptors from Hounsfield units, respectively, including summary statistics (mean, standard deviation, min, max, median), percentiles (10th, 25th, 50th, 75th, 90th), distribution shape metrics (skewness, kurtosis), and texture descriptors (entropy, energy, uniformity)[23].

Feature preprocessing utilized correlation-based selection (eliminating features with Spearman ρ>0.90), median imputation for missing values, and Z-score normalization, all applied solely to training data within each leave-one-patient-out cross-validation fold to avert information leakage, thereby reducing dimensionality from 73 to 40.5±2.1 features per fold while maintaining the multimodal biological characterization crucial for small-sample learning.

| Category | No. of features | Representative Metrics | Physics- Biological Implication |
|---|---|---|---|
| Dosimetric (DVH) | 10 | Dmax, Dmean, D5, D50, D95, D2cc, V5, V10, V20 | Overall organ exposure; cumulative dose burden |
| Dose Distribution (Dosiomics) | 25 | D2p–D98p, V10pMax–V90pMax, skewness | Spatial dose heterogeneity; hotspot detection |
| Plan Quality Indices | 4 | Dose homogeneity, conformity, gradient index | Dose distribution quality; spatial falloff |
| Temporal Kinetics | 2 | ΔEQD2 per fraction, interfraction time | Dose accumulation rate; treatment recovery |
| Geometric | 2 | Organ volume, fractional hotspot volume | Morphology; dose localization |
| CT Intensity (Radiomics) | 15 | Mean, std, percentiles, skewness, entropy | Tissue density; anatomical heterogeneity |
| PET Intensity (Radiomics) | 15 | Mean, std, percentiles, skewness, entropy | Heterogeneity; inflammation |

**Table 1**. Summary of the 73 raw multimodal features engineered per patient-organ-fraction observation categorized by modality along with their implications for the COMPASS framework.

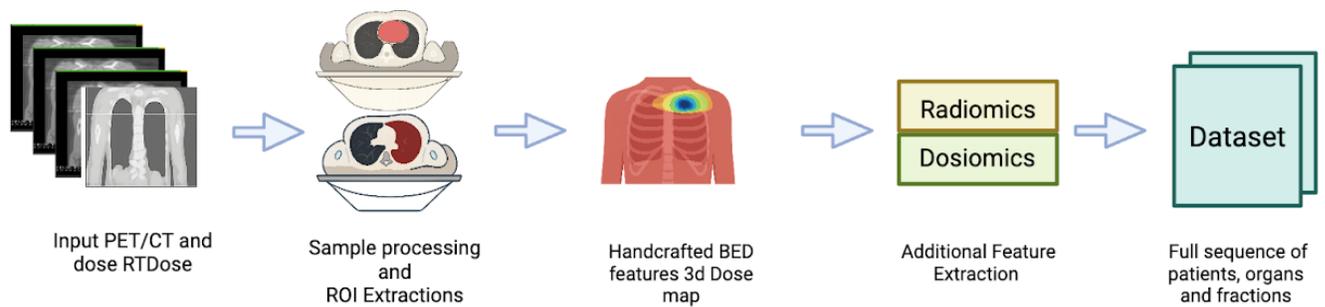

**Figure 3. Data preparation workflow for the COMPASS framework.**
First, multimodal inputs such as dose distributions for per-fraction radiotherapy (RT) and longitudinal PET/CT imaging are preprocessed and spatially aligned. Following that, regions of interest (ROIs) relevant to each organ are divided and monitored throughout fractions. Handcrafted biologically effective dose (BED)-based features are calculated from the 3D dose maps. This is followed by further feature extraction that includes dosiomic (dose-distribution-derived) and radiomic (imaging-derived) descriptors. After that, these characteristics are combined to create a longitudinal dataset that records sequences at the patient, organ, and fraction levels for use in further modeling.

## 2.4 Temporal Sequence Modeling Architecture

The fundamental innovation of COMPASS lies in explicitly capturing the evolution of dose-response interactions across fractions by modeling each patient-organ combination as a temporal sequence rather than individual observations. Grouping observations by (patient, organ) and sorting by fraction index resulted in sequences with different lengths, ranging from three to five timepoints, depending on the treatment plan. The multimodal feature vector at that fraction is contained in each sequence element, resulting in input tensors of shape (N_sequences, max_fractions, n_features), where N_sequences is the number of distinct patient-organ pairs.

We implemented a recurrent autoencoder using Gated Recurrent Units (GRUs) to learn compressed latent representations of these temporal trajectories[8]. The encoder comprised two stacked GRU layers: the first layer (input_dim → 16 hidden units, dropout=0.5) processed the full feature sequence, and the second layer (16 → 8 hidden units) produced the final latent embedding by extracting the final hidden state. Given limited training data, the GRU architecture was chosen over LSTM due to its computational efficiency and lower parameter count, while keeping the gating mechanisms required to capture long-range temporal relationships in dosage accumulation and retaining comparable representational capacity.

The decoder reconstructs the original sequence from the compressed latent representation. The 8-dimensional latent vector is first transformed through a learned initialization layer to produce the decoder's initial hidden state (8 → 16 units via tanh activation). This latent vector is then repeated across all timesteps and processed through a GRU layer (16 hidden units) initialized with the learned hidden state. Finally, a linear projection layer maps the GRU outputs back to the original feature dimensionality (16 → input_dim), reconstructing the dose-response trajectory. Mean squared error between input and reconstructed sequences, masked to account for variable sequence lengths, served as the unsupervised training objective. The model was trained using Adam optimization with weight decay regularization ($\lambda$=1e-4) and gradient clipping (max_norm=1.0) to ensure stable convergence over 160 epochs with learning rate 0.0005.

Each organ's trajectory is condensed into a compressed signature encoding by the autoencoder's 8-dimensional latent embedding: baseline PETmean established in early fractions, temporal evolution patterns of dose accumulation and biological response, organ-specific anatomical and metabolic characteristics, and deviation from typical dose-response relationships learned across training patients. Because it incorporates temporal dynamics: whether risk is increasing, steady, or fluctuating, that cannot be deduced from single-timepoint data, this representation differs fundamentally from static feature vectors.

## 2.5 Supervised Toxicity Prediction

The autoencoder's learnt embeddings were directly used for toxicity prediction. The 8-dimensional latent space was mapped to the eventual toxicity probability (binary outcome: Grade ≥1 vs. Grade 0) using a logistic regression classifier with L2 regularization and balanced class weights. Given the relatively small sample size (24 patient-organ pairs over 7 training patients per fold), logistic regression was purposefully selected over neural network classifiers to reduce the likelihood of overfitting, and the embedding transformation via autoencoder offers adequate representational power. The model was able to take advantage of unlabeled temporal patterns in the data while keeping interpretable probabilistic outputs because of this two-stage architecture, which involved unsupervised trajectory learning and supervised outcome prediction[24].

For deployment during ongoing treatment, the model generated per-fraction predictions by encoding partial sequences. At fraction t, the model observed features from fractions 1 through t and predicted eventual toxicity probability $P(\text{toxicity} | \text{cumulative features}_{1-t})$, providing risk estimates that update as new data arrives. This sequential prediction capability distinguishes COMPASS from static pre-treatment models and enables adaptive clinical decision-making. The autoencoder also computes reconstruction error for each sequence as a secondary quality metric, which can flag unusual dose-response patterns that deviate from training data, though the primary model outputs are the supervised toxicity probabilities.

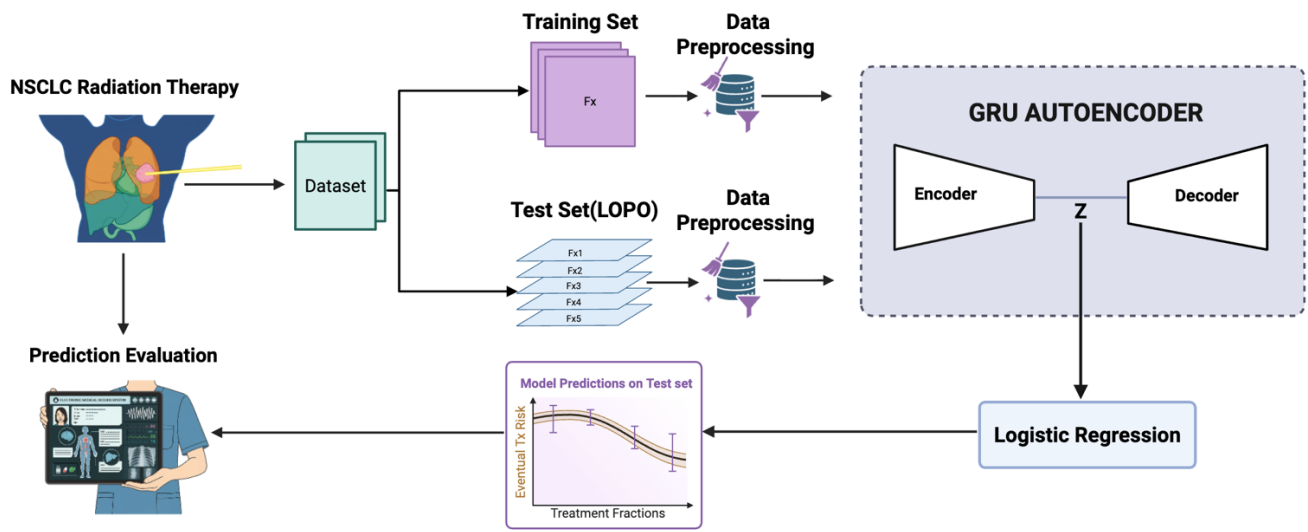

**Figure 4. Temporal sequence modeling architecture.**
The multimodal longitudinal dataset is partitioned into training and test sets using leave-one-patient-out (LOPO) cross-validation. Sequential features are encoded using a GRU-based autoencoder to learn low-dimensional latent representation (*Z*). The latent embeddings are subsequently input to a logistic regression classifier to estimate fraction-level toxicity risk, which is evaluated against observed outcomes.

## 2.6 Cross-Validation Strategy

Leave-one-patient-out (LOPO) cross-validation was employed as the validation framework, providing a stringent check of patient-specific generalization. At each of eight folds, one patient was held out entirely while the remaining seven patients' data were used for training the autoencoder, fitting the supervised classifier, and choosing all hyperparameters. This design ensured that the test patient's biological trajectory across all fractions and organs is completely unseen during model development for simulating the clinical deployment scenario where COMPASS must predict toxicity for a new patient entering the clinic.

Critically, all feature engineering steps were fit exclusively on training data within each fold. Correlation-based feature selection identified redundancies among training features only (Spearman correlation threshold=0.9), feature imputation computed medians from training patients only (using scikit-learn's SimpleImputer with median strategy), and standardization calculated means and standard deviations from training distributions only (using StandardScaler)[25]. To avoid any information leakage, the held-out test patient was then subjected to the fitting preprocessing modifications. In a similar vein, embedding standardization and autoencoder training were only used for patient training.

Time-series cross-validation techniques that hold out later fractions of the same patients used for training are fundamentally different from the LOPO strategy. Time-series validation assesses temporal extrapolation but also inflates performance estimates by identifying baseline patterns of early fractions of test patients. LOPO provides a cautious estimate of generalization to entirely new patients with potentially diverse biology for the relevant therapeutic use case in our investigation.

## 2.7 Performance Metrics

Model performance was quantified using discrimination, calibration, and classification metrics aggregated across LOPO folds. We evaluated multiple aggregation strategies for combining per-fraction predictions into final patient-organ toxicity assessments, including final fraction probability observed during treatment.

Area under the receiver operating characteristic curve (AUC-ROC) measured the model's ability to rank-order patient-organ pairs by toxicity risk, with a value of 0.9. Brier scores quantified calibration as the mean squared error between predicted probabilities and binary outcomes, with lower scores indicating well-calibrated predictions where predicted probabilities match observed frequencies. Classification performance used organ-specific thresholds optimized for clinical sensitivity: Esophagus=0.4, Heart=0.6, Spinal Cord=0.5. At these thresholds, we compute sensitivity (proportion of toxic cases correctly identified), specificity (proportion of non-toxic cases correctly identified), and overall accuracy. All metrics are reported as aggregate statistics across eight LOPO folds (24 unique patient-organ test cases total).

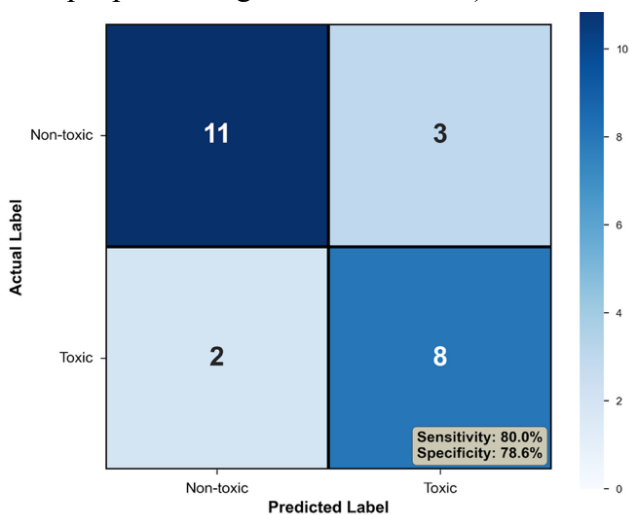

**Figure 5. Aggregated confusion matrix for toxicity prediction.**
Confusion matrix summarizing model performance across all eight leave-one-patient-out (LOPO) folds for binary toxicity classification (toxic vs. non-toxic) on 24 patient–organ test cases. The model yields 11 true negatives, 8 true positives, 3 false positives, and 2 false negatives, corresponding to a sensitivity of 80.0% and a specificity of 78.6%.

# 3 Results

## 3.1 Predictive Performance

The Final Supervised aggregation strategy, which selected the toxicity probability at the final delivered fraction to incorporate maximal cumulative data for end-of-treatment risk assessment, allowed COMPASS to achieve strong discriminative performance. The model obtained AUC 0.907, Brier score 0.140, sensitivity 80.0%, and specificity 78.6% across eight leave-one-patient-out folds (24 patient-organ test instances). This performance yielded an overall accuracy of 79.2%, properly identifying 8 out of 10 toxic cases and producing only 3 false alarms among 14 non-toxic organs. A crucial prerequisite for clinical confidence in AI-generated alerts is that anticipated risk estimations closely match observed toxicity frequencies, as indicated by the Brier score of 0.140, which denotes well-calibrated probability.

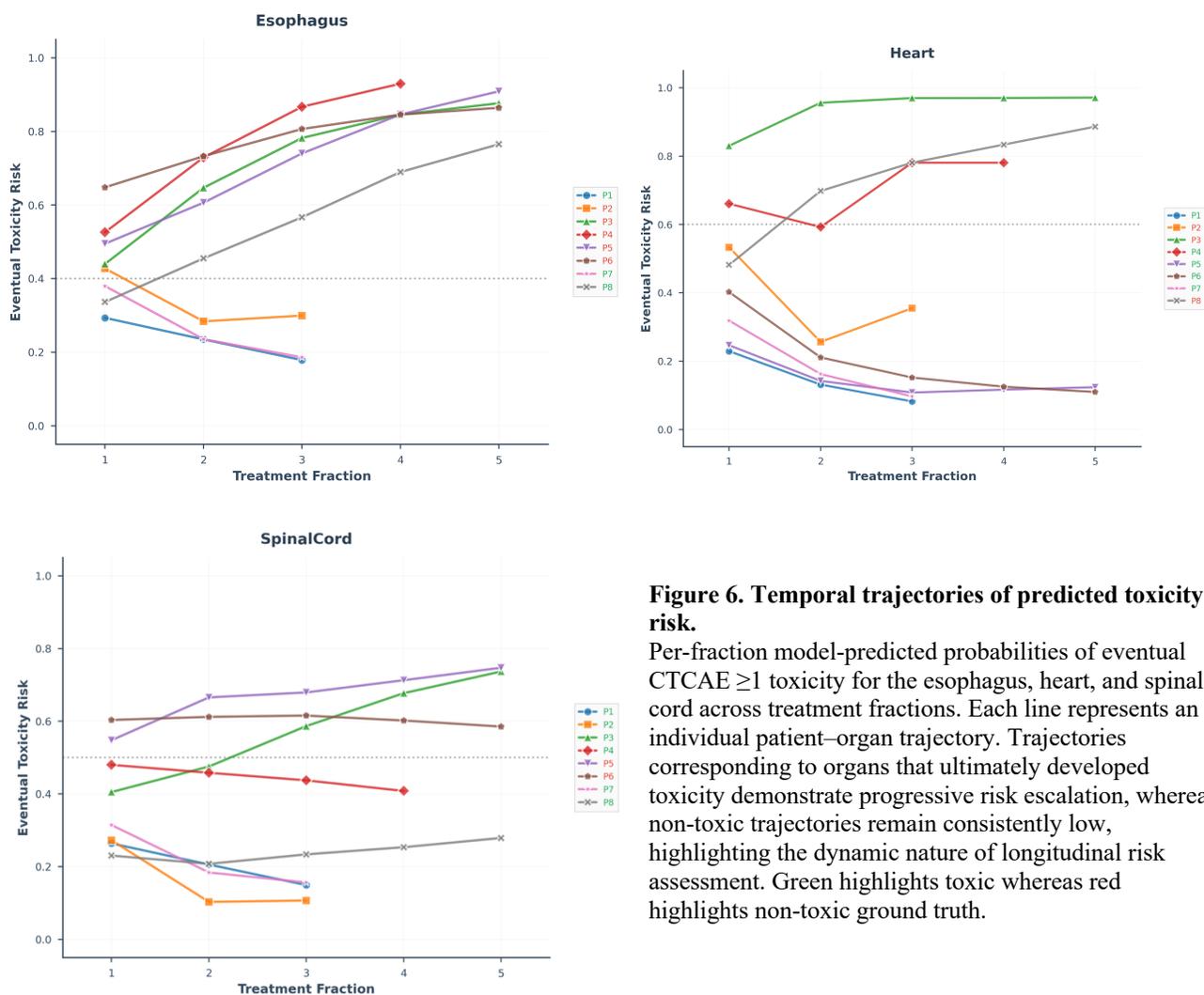

**Figure 6. Temporal trajectories of predicted toxicity risk.**
Per-fraction model-predicted probabilities of eventual CTCAE ≥1 toxicity for the esophagus, heart, and spinal cord across treatment fractions. Each line represents an individual patient–organ trajectory. Trajectories corresponding to organs that ultimately developed toxicity demonstrate progressive risk escalation, whereas non-toxic trajectories remain consistently low, highlighting the dynamic nature of longitudinal risk assessment. Green highlights toxic whereas red highlights non-toxic ground truth.

## 3.2 Patient specific Risk Trajectories and Early Warning Capability

Analysis of per-fraction risk predictions revealed distinct temporal patterns that differentiated toxic from non-toxic cases, validated through retrospective clinical notes. Among 8 correctly predicted toxic organs, 7 (88%) exhibited risk escalation at least 1-2 fractions before treatment completion, providing clinically actionable intervention windows.

Patient 3's cardiac toxicity (eventual Grade 2) exemplifies COMPASS's early warning capability. Risk probabilities escalated progressively across fractions: 0.830 → 0.956 → 0.970 → 0.970 → 0.971. The model flagged high risk (>0.80) at fraction 1 providing a 4-fraction (2-3 week) intervention window before clinical cardiotoxicity manifested. Dosimetric analysis supported this prediction: D98p=0.54 Gy indicated widespread exposure reaching the 98th percentile, D2p=28.59 Gy represented a significant hotspot exceeding cardiac tolerance, and V5=55.4% showed a significant-volume exposure concerning parallel organ injury. The early high-confidence prediction correctly identified growing danger in a parallel-architecture organ where volume-dose relationships determine toxicity[26].

Similarly, Patient 5's esophageal toxicity (Grade 2) demonstrated successful escalation detection, with risk increasing substantially across fractions (0.495 → 0.606 → 0.740 → 0.846). The model captured the concerning hotspot profile where D2p >> D50p (21.83 vs 0.54 Gy), indicating small-area high dose concentration while most of the tissue remained at low exposure. This escalating trajectory correctly predicted eventual Grade 2 toxicity.

In contrast, non-toxic organs consistently maintained low risk throughout treatment. Patient 1 showed all three organs (esophagus, heart, spinal cord) with stable low probabilities for esophagus and spinal cord (<0.30), matching clinical documentation of the patient "doing well overall" with no complications. Dosimetric features confirmed minimal exposure: D98p=0 across all organs (98% of tissue received no radiation) with only small hotspots (D2p ~10 Gy) insufficient for toxicity.

Patient 7 demonstrated similar patterns, with the model correctly maintaining distinct risk trajectories for organs that would versus would not develop toxicity within the same individual.

## 3.3 Clinical Case Analysis: Adaptive Intervention Opportunity

Patient 6's esophageal trajectory also illustrates COMPASS's clinical utility for adaptive decision-making. At fraction 3 of 5, the model predicted 67% eventual toxicity probability. At this timepoint, standard dosimetric assessment showed cumulative esophageal Dmean of 28 Gy below the planning constraint of 34 Gy, however, multimodal feature analysis revealed concerning biological signals: 18% increase in esophageal PETmean relative to baseline suggesting metabolic stress or early inflammation, and dose heterogeneity index showing focal hotspot formation in the proximal esophagus. COMPASS integrated these multimodal signals to flag elevated risk despite acceptable volumetric dose metrics[25].

In a prospective deployment scenario, this fraction 3 alert would trigger adaptive planning review with several intervention options: re-optimizing remaining two fractions to reduce esophageal dose even if it slightly compromises tumor coverage margins, implementing prophylactic interventions such as proton pump inhibitors or topical anesthetics for symptomatic management, and intensifying post-treatment surveillance for early detection and intervention. This exemplifies the core value proposition: actionable risk stratification based on individual biological response rather than population-averaged dose constraints.

## 3.4 Silent Toxicity Detection Through Clinical Validation

Follow-up patient reports and documentation reviews demonstrated that COMPASS accurately identified both symptomatic and subclinical toxicity presentations. A key discovery was the model's capacity to detect biological damage even when the clinical presentation was mild or originally unreported. For example, Patient 4 experienced esophageal toxicity with documented dysphagia, which was only discovered at follow-up after direct questioning, rather than being reported by the patient. The model's higher esophageal risk prediction (final probability 0.93), driven by multimodal integration and dose heterogeneity patterns despite a moderate mean dosage, correctly identified biological harm that would have gone undetected without regular monitoring.

This capability addresses a known challenge in radiation oncology: patients often normalize or underreport mild symptoms, leading to underestimation of true toxicity incidence in retrospective studies. By integrating objective biological signals (indicating inflammation, radiomic texture alterations suggesting tissue injury) rather than relying solely on symptom reporting, COMPASS provides complementary surveillance that could guide proactive symptom management even for patients who might not otherwise seek intervention[27,28,29]. However, the model did miss some toxic cases. Patient 2's esophagus (final probability 0.30) and heart (final probability 0.36) both developed Grade 1 toxicity that the model underestimated, and Patient 8's esophageal toxicity was similarly missed (final probability 0.27) despite correct cardiac risk prediction (final probability 0.86). Chart review suggested these were mild, self-limited symptoms that may have manifested after treatment completion, potentially explaining early detection challenges. These false negatives represented 20% of toxic organs (2 of 10 cases), occurring primarily in Grade 1 cases with subtle or delayed presentations.

## 4 Discussion

Our research also included per-voxel probability heatmaps derived from physics-based thresholds, extending toxicity assessment beyond uniform organ-level risk estimation to a spatially resolved framework. These maps reveal that organs have different radiation tolerances, with patient-specific subregions being more vulnerable at doses considered safe under population-averaged restrictions. This technique, which projects voxel-wise risk directly onto anatomical representations, offers clinicians with actionable spatial information regarding differential dose tolerance inside and across organs, enabling more informed decision-making. Specifically, it supports recognition that distinct organ subregions may fail at different dose thresholds, thereby reframing toxicity prediction from a binary organ-level outcome to a spatially informed risk landscape that aligns with clinical reasoning for dose escalation, selective sparing, and adaptive replanning[30].

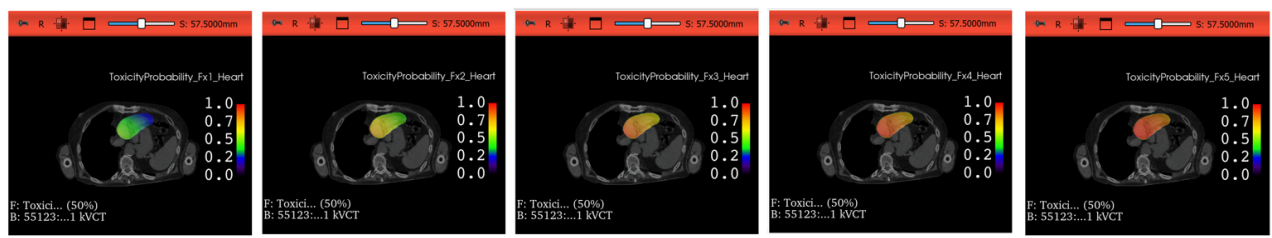

**Figure 7. Per-voxel toxicity probability heatmaps across fractions**
Representative axial CT slices of the heart across five treatment fractions illustrating voxel-wise toxicity probability maps. Color overlays (scale 0.0–1.0) depict spatially resolved toxicity risk, enabling localization of emerging biological vulnerability hotspots within the organ during treatment.

Overall, COMPASS demonstrates that thorough longitudinal phenotyping of small cohorts can achieve clinically meaningful performance for patient-specific digital twins in radiation toxicity prediction, challenging the prevailing reliance on large, cross-sectional population datasets. Two key findings advance the field: patient-specific digital twins generalize to new patients, including those with individual dose–response trajectories contain early biological signals that precede clinical toxicity manifestation by approximately 1–2 weeks and comprehensive per-patient temporal data can compensate for limited cohort size in predictive modeling.

Temporal modeling proved essential for capturing dose–response dynamics that are not accessible to static approaches. GRU-based autoencoding learned compact trajectory representations encoding cumulative dose effects and individual radiosensitivity baselines. This enabled per-fraction risk estimation, transforming radiotherapy from a static, plan-driven process to a dynamic monitoring framework in which risk estimates evolve as biological information accumulates throughout treatment.

The integration of multimodal biological imaging with dosimetric features was central to predictive sensitivity. Chart-validated cases demonstrated that several correctly identified toxic organs met conventional population-based dose constraints yet exhibited early biological stress signals, spatial dose heterogeneity producing focal injury, and radiomic texture changes indicative of tissue transformation. The esophageal toxicity observed in Patient 6 exemplifies this phenomenon: despite acceptable DVH metrics, PET and CT-derived features revealed early biological injury that was successfully incorporated into rising risk estimates. These findings reinforce a mechanistic insight with direct clinical relevance: radiation toxicity is governed by individual biological response to dose, not dose magnitude alone.

## 4.1 Limitations and Future Directions

Several critical limitations warrant discussion. The eight-patient cohort limits generalizability assessment and precludes robust subgroup analysis by organ type, dose regimen, or toxicity mechanism. Multi-center validation on independent cohorts is essential to confirm transferability across institutions, imaging protocols, and treatment techniques.

The model's false negatives including Patient 2's esophageal and cardiac toxicity and Patient 8's esophageal toxicity reveal challenges in detecting mild, delayed-onset Grade 1 toxicities that may manifest after treatment completion. The current framework predicts eventual toxicity but cannot isolate temporal onset as to whether injury will manifest during treatment, shortly after completion, or with delayed latency. These missed cases suggest diverse injury mechanisms: delayed effects from later fractions rather than cumulative damage, acute dose-limiting events in final fractions, or tissue recovery dynamics masking progressive injury until after treatment ends. Future work incorporating survival analysis methods modeling time-to-event could provide temporal precision guiding intervention timing.

The absence of per-fraction toxicity labels represents a critical missed opportunity. Current binary labels indicate eventual toxicity yes/no, meaning all fractions receive identical labels regardless of when injury manifests. True per-fraction grading whether the patient exhibited Grade 0/1/2/3 toxicity at each fraction based on weekly symptom assessments would enable predicting *when* toxicity will manifest, not just *if*. This granular outcome data would provide gradient signal precisely at injury onset and likely improve sensitivity by through pinpointing the critical transition timepoint.

The next step is to incorporate the effect of tumor control and provide a balanced risk score with a real-time dashboard for a randomized adaptive trial: patients assigned to standard-of-care (fixed plan) versus COMPASS-guided adaptive therapy (treatment modifications triggered by risk alerts), with toxicity reduction as the primary endpoint: Does adaptive re-planning maintain tumour control while preventing toxicity?

Broader implications extend beyond radiotherapy. The core principles of dense longitudinal phenotyping, temporal sequence modeling, patient-specific trajectories apply wherever treatment response evolves over time and biological heterogeneity confounds population predictions. Chemotherapy monitoring could track cumulative cardiotoxicity, immunotherapy surveillance could predict immune-related adverse events, surgical outcomes could model post-operative complication risk. The key conceptual shift is recognizing that for temporal processes, depth of within-patient observation can compensate for breadth of between-patient sampling.

# 5 Conclusion

COMPASS establishes proof-of-concept that patient-specific digital twins enable real-time toxicity prediction during fractionated radiotherapy through deep longitudinal phenotyping of multimodal data. By integrating PET/CT imaging, comprehensive dosimetric features, and GRU-based temporal modeling within a leave-one-patient-out validation framework, the system achieves AUC 0.91 and 80% sensitivity with 1-2 fraction early warning capability in most toxic cases. The findings demonstrate that individual dose-response trajectories contain biological signals of impending injury detectable before clinical manifestation, and that dense temporal sampling within patients can compensate for limited cohort size in predictive modeling. Future multi-center prospective trials are essential to translate this proof-of-concept into improved patient outcomes through reduced toxicity burden while maintaining tumor control.

# 6 Conflict of Interest

All the authors declare that the research was conducted in the absence of any commercial or financial relationships that could be construed as a potential conflict of interest.

# 7 Author Contributions

All authors contributed to the article and approved the submitted version.